# Transfer Learning and Meta Classification Based Deep Churn Prediction System for Telecom Industry

Uzair Ahmed[1], Asifullah Khan[1,2], Saddam Hussain Khan[1], Abdul Basit[3], Irfan Ul Haq[1], and Yeon Soo Lee[4]*

[1]Pattern Recognition Lab, Department of Computer & Information Sciences, Pakistan Institute of Engineering & Applied Sciences, Nilore, Islamabad 45650, Pakistan

[2]Center for Mathematical Sciences, Pakistan Institute of Engineering & Applied Sciences, Nilore, Islamabad 45650, Pakistan

[3]TPD, Pakistan Institute of Nuclear Science and Technology, Nilore, Islamabad, Pakistan

[4]Department of Biomedical Engineering, College of Medical Science, Catholic University of Daegu, South Korea

asif@pieas.edu.pk

Abstract: A churn prediction system guides telecom service providers to reduce revenue loss. However, the development of a churn prediction system for a telecom industry is a challenging task, mainly due to the large size of the data, high dimensional features, and imbalanced distribution of the data. In this paper, we present a solution to the inherent problems of churn prediction, using the concept of Transfer Learning (TL) and Ensemble-based Meta-Classification. The proposed method "TL-DeepE" is applied in two stages. The first stage employs TL by fine-tuning multiple pre-trained Deep Convolution Neural Networks (CNNs). Telecom datasets are normally in vector form, which is converted into 2D images because Deep CNNs have high learning capacity on images. In the second stage, predictions from these Deep CNNs are appended to the original feature vector and thus are used to build a final feature vector for the high-level Genetic Programming (GP) and AdaBoost based ensemble classifier. Thus, the experiments are conducted using various CNNs as base classifiers and the GP-AdaBoost as a meta-classifier. By using 10-fold cross-validation, the performance of the proposed TL-DeepE system is compared with existing techniques, for two standard telecommunication datasets; Orange and Cell2cell. Performing experiments on Orange and Cell2cell datasets, the prediction accuracy obtained was 75.4% and 68.2%, while the area under the curve was 0.83 and 0.74, respectively.

1. Introduction

'Customer churn' is a term related to the customer subscription model and refers to the process, whereby a customer terminates to use the services of a provider [1]. This term is frequently used in the telecommunication industries [2, 3], since several factors like coverage area and lucrative offers from competitors can trigger a user's decision to switch from one service provider to another. Telecom corporations can benefit greatly by retaining customers that are on the verge of churning. This is because it has been estimated that the cost of retention is usually 5 to 15 times less as compared with the cost of acquiring a new customer [4, 5]. The corporations, therefore, need an effective churn prediction system to automate the task of detecting the customers who might churn. The prediction is usually made on customer's usage patterns and other factors, which directly or indirectly affect the user's decisions.

Customer Churn Prediction is a complex problem having challenges such as messy data, low churn rate, churn event censorship, etc. [6]. The current customer churn prediction methods are largely based on machine learning classification methods. However, the performance of a classifier generally suffers due to high dimensions of the telecommunication datasets. Furthermore, the telecommunication data has an imbalanced nature in its distribution (with a limited number of examples of the minority class) that also hampers in achieving accurate churn prediction. Churner prediction is considered as a binary classification problem, where churners lie in the minority class and non-churners lie in the majority class. Several machine learning algorithms focus on improving the whole classification performance by sacrificing the accurate prediction of churners in the minority classes. Generally, churners (minority class samples) are more misclassified as compared to non-churners, due to the fundamental problem of its imbalanced nature. Due to the reason, machine learning based classification has a tendency to classify all samples as non-churners, which can provide good accuracy but less precision in terms of predicting churner's class. These complications in churn prediction show that there is still a margin of improvement in performance, keeping in view the new challenges and opportunities that are arising in the domain of big data analytics.

Previously, Support Vector Machines (SVMs) were applied to improve the performance of telecommunication churn prediction. Similarly, Filter and Wrapper methods were collectively used for feature selection, and their selected features were further provided to ensemble classifiers. These ensemble classifiers included Random Forest, RotBoost and SVMs [7, 8] that explore the selected feature vectors helpful for churn classification [9]. Yabas et al. presented meta-classification methods [10] for standard telecom Orange dataset. Their meta-classification methods include bagging, Random Forest, Logistic Regression and Decision Trees, which reported an Area Under the Curve (AUC) of 0.7230. Along similar lines, Koen et al. applied Rotation Forest and AdaBoost-based ensemble and reported an AUC as 0.697 [11]. Verbeke et al. have proposed different methods including SVMs, Naïve Bayes, k-NN, Neural Network and reported AUC of 0.714 on standard telecommunication Orange dataset [12]. Similarly, a work using an ensemble classifier with mRMR based processing [13] reported that RotBoost and Random Forest-based ensemble [14] resulted in an AUC of 0.749. The approach of GP and AdaBoost ensemble. The approach of GP and AdaBoost ensemble [15, 16] was employed for churn prediction telecommunication data and reported an AUC of 0.63.

On the other hand, different strategies have been proposed to overcome class imbalance problems, which are divided into two main categories; Algorithm Level and Data Level. Nowadays, Transfer Learning method has also been used for prediction problems with imbalanced nature of dataset. The motivation of employing Transfer Learning in conjunction with ensemble algorithms for telecommunication churn prediction data classification is three-fold. Firstly, several ensemble algorithms were applied by researchers and mostly did not show satisfactory classification performance, i.e. only showed marginal improvement in the churn prediction. Secondly, Convolution Neural Networks (CNNs) are known to achieve better performance on images or grids like input topology. Consequently, in this work, feature vectors are converted into images, and deep learning methods are applied to achieve better performance. Finally, it has been observed through various research studies that Transfer Learning [17-20] using fine-tuning of an already trained network may give an improved performance as compared with training a CNN from scratch.

The proposed Transfer Learning and Ensemble-based Meta-Classification "TL-DeepE" exploits three ideas to develop an effective prediction system:
1. Firstly, it uses Deep CNNs as base classifiers for the ensemble classification. Moreover, both original and converted features are considered. The original numerical features are converted into 2D image data first and then, are provided to the base classifiers.

2. Secondly, the proposed methodology exploits the quick learning and knowledge transfer abilities of "Transfer Learning". To perform Transfer Learning in the CNNs models, TF Slim library [21] is used.

3. Thirdly, we exploit the learning and discrimination abilities of the high-level Genetic Programming and AdaBoost based ensemble classifier. In this regard, the predictions of the Deep CNNs are appended to the original feature vector and thus are used to build a final feature vector for the high-level ensemble classifier.

The remaining sections of this manuscript are organized as follows: Section 2 provides details of the proposed TL-DeepE technique. It explains both the Transfer Learning idea as well as the GP-AdaBoost ensemble-based classification. Section 3 provides details of the telecommunication dataset and experimental setup used in this work. Results and comparative analysis are discussed in section 4. Finally, section 5 concludes the paper.

2. Proposed TL-DeepE Prediction System

The proposed TL-DeepE technique for telecom churn prediction uses the concept of meta-classification in addition to Transfer Learning (Figure 1). Input features are converted from one-dimensional vector to two-dimensional image format and are used to train multiple CNNs, which in turns give churn predictions (collectively called "prediction space"). These predictions are then appended with the original feature vector, and classification is performed on this "extended" feature set using GP-AdaBoost ensemble. Experimental evaluations are performed by using 10-fold cross-validation on two standard telecommunication churn prediction datasets i.e. from Orange and Cell2cell. The exploitation of sample space by AdaBoosting and enhancement of the learning and generalization by GP [22] makes the proposed TL-DeepE an effective method for churn prediction and classification.

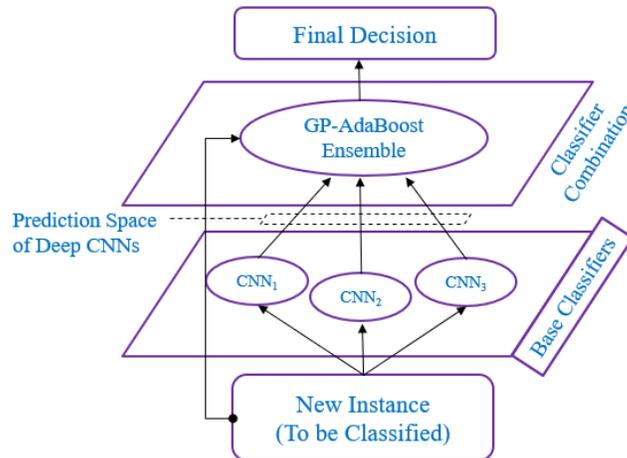

Figure 1: An Overview of the proposed TL-DeepE Technique.

2.1  Data Pre-Processing

Telecom datasets have many concerns to be addressed like missing values, non-numeric feature values, inconsistent feature scales, etc. It is, therefore, necessary to pre-process the data before applying a learning model. The datasets used, namely Cell2cell and Orange, have their individual characteristics. Initially, missing values from the dataset were handled by numeric and categorical feature vectors, respectively. Missing values from the dataset were inserted as column mean and mode for numeric and categorical feature vectors. This was done for the features which had not an overwhelming number of missing values. The features with more than 95% of missing values were removed from the dataset since it was assumed that they contribute little to the learning. Next, the categorical features were converted to numeric using the 'dummy variable' technique, where different categories are split to form their own features. The value of 1 or 0 is assigned, depending on whether the instance had the category as its original feature or not, respectively. Once the entire dataset was converted into a numeric format, we normalized it on a scale of 0 to 1 in order to scale the features uniformly.

## 2.2 Conversion of 1D Features to Images

A major aspect of this research work was to explore the possibility of converting the raw feature vector-based dataset to image form, with the belief that enhanced performance can be achieved this way using CNNs. The assumption behind this hypothesis is that CNNs are known to perform well on image data. Every single instance present in the dataset was converted to an individual representative image. To achieve this, the one-dimensional feature vector of each instance was converted into a two-dimensional matrix form. This conversion was performed on the hit-and-trial basis, and it was found that the images obtained using the method of interpolation works well as far as the accuracy is concerned. Specifically, we constructed a 2-dimensional matrix of normalized pixels, where each row is for each day and each column is for each type of behavior. This conversion method was used to convert all instances to image format and in turn, (both in case training and testing). It is hypothesized that one can exploit the pixel-based differences between churner and non-churner instance images when it comes to CNNs learning.

## 2.3 Development of Individual Prediction Spaces using Transfer Learning

Training a CNN on a new dataset from scratch can be a very time consuming and resource intensive practice. The concept of Transfer Learning offers an alternative route to avoid learning from scratch. This concept introduces that instead of training from scratch, CNNs already trained on massive dataset like ImageNet should be fine-tuned (i.e. the weight updatation is performed) on the target dataset.

In order to obtain the decision spaces, three CNN models were used: AlexNet, Inception-ResNet-V2, and a custom 6-layer neural network model. These CNNs were pre-trained on ImageNet dataset and Transfer Learning was used to fine-tune them on the Orange and Cell2cell datasets.

### 2.3.1 AlexNet

AlexNet is a famous CNN architecture proposed by Alex Krizhevsky for ImageNet classification [23-25]. The architecture is composed of eight layers: first five are convolution layers with 4-pixel stride and the next three are fully-connected and lead to the final layer, i.e. softmax. Softmax layer provides the probabilistic distributed decision over 1000 classes. The multinomial logistic regression-based objective is maximized by the network. The layers have kernels smaller in size but an increased number of channels, such as 5x5x48 and 3x3x256. Fully-connected layers each have 4096 neurons and a final 1000-way softmax layer is employed at the end. The network uses data augmentation and dropout to reduce the effects of overfitting.

### 2.3.2 Inception-ResNet-V2

Inception family is a class of deep CNN architectures developed by the Google team. Inception family architectures usually need more computational power as compared with AlexNet because of their more complex architecture. The concept behind Inception-ResNet-v2 [26] was to train deeper neural network architectures using skip connections [27, 28] and to automate the filter size selection. A single block in Inception ResNet architecture uses a filter bank with multiple sizes of filters (5x5, 3x3, 1x1) On the ILSVRC image classification benchmark, this architecture demonstrated an accuracy of 95.3%, which was among the top 5 of its categories [29].

### 2.3.3 Custom-CNN Architecture

For comparison purposes, a custom deep neural network [30, 31] architecture was also trained after hit-and-trial experimentation with different architectural parameters. The architecture was built in the layer patterns of conv-relu-pool-fc-softmax. It takes 32x32 input image, has a 3x3x6 filter with no padding at first Convolution layer and then 2x3 pooling, 5x5x10 conv filter in the second layer, 2x2 max-pooling, and the final classification layer consists of a softmax with 2 neurons.

These custom-CNNs were originally pre-trained on ImageNet data and then were fine-tuned on the Orange and Cell2cell datasets using Transfer Learning. Once fine-tuned, the models were tested on the test set of images and their predictions were collected. These predictions were then appended to the original features to obtain extended feature space. The extended feature space was passed on for training and testing of the high-level GP-AdaBoost ensemble classifier.

### 2.3.4 Exploitation of Individual Prediction Spaces using GP-AdaBoost Ensemble Based Meta Classification

GP has been reported to be quite successful in solving problems related to classification, clustering, and regression. In this work, GP concentrates on developing effective one-class classifiers to come up with an improved prediction space. The Adaboost, on the other hand, tries to provide stringent sample space to the one-class classifiers by intensifying the hard examples. Finally, for an overall binary decision, a maximum weighted strategy is applied. GP-AdaBoost ensemble [32] incorporates the idea of boosting in evolving different GP programs. A classifier in a subsequent step is evolved for each class to identify the "hard samples" that were incorrectly classified by the preceding classifier. The method in an earlier stage, divides the data into two sets, i.e. training and testing.

In Figure 2, a number of GP programs are evolved for a given fixed Elite size in each class, and then, weighted sum is computed for each class. A class with the highest weighted sum is then predicted by the GP-Adaboost. Final prediction is made using the highest weighted sum output in each class.

The GP programs are basically one-class classifiers and therefore, evolved separately for both churners and non-churners. In this work, the Genetic Programming-AdaBoost [33] ensemble is used as a high-level meta-classifier on top of the base results achieved by the CNNs and Transfer Learning. The proposed method extracts discriminative features using deep autoencoder and concatenates it to the original feature vector to form images. The extended feature vector uses the original feature vector concatenated with the base CNNs predicting space as input and outputs the final predictions.

3. Performance Evaluation

Generally, results of churn prediction systems are compared by standard measures like prediction accuracy. But in telecommunication datasets, accuracy is not sufficient to achieve the true picture of results due to an imbalanced distribution of the data. Therefore, the Area Under Curve (AUC) of Receiver Operating Characteristics (ROC) curve along with prediction accuracy is used as a performance measure to estimate the true functionality of the predictor. ROC [34] curve are generated on the basis of true positive rate (Sensitivity) and false negative rate (1-Specificity). Let TP denotes True Positives, FP False Positives, TN True Negatives, and FN False Negatives. The positive prediction (TP+FN) shows the true churners rate and negatives prediction (TN+FP) shows non-churners rate.

3.1 Prediction Accuracy

Prediction accuracy (Equation 1) gives a measure of the correct predictions against the total number of cases evaluated. The metric of prediction accuracy has been considered for the Transfer Learning task since the number of correct predictions from the neural network models would affect the overall ensemble classifier's result. The prediction accuracy is defined as below:

$$Prediction\ Accuracy = \frac{TP+TN}{TP+TN+FP+FN} \quad (1)$$

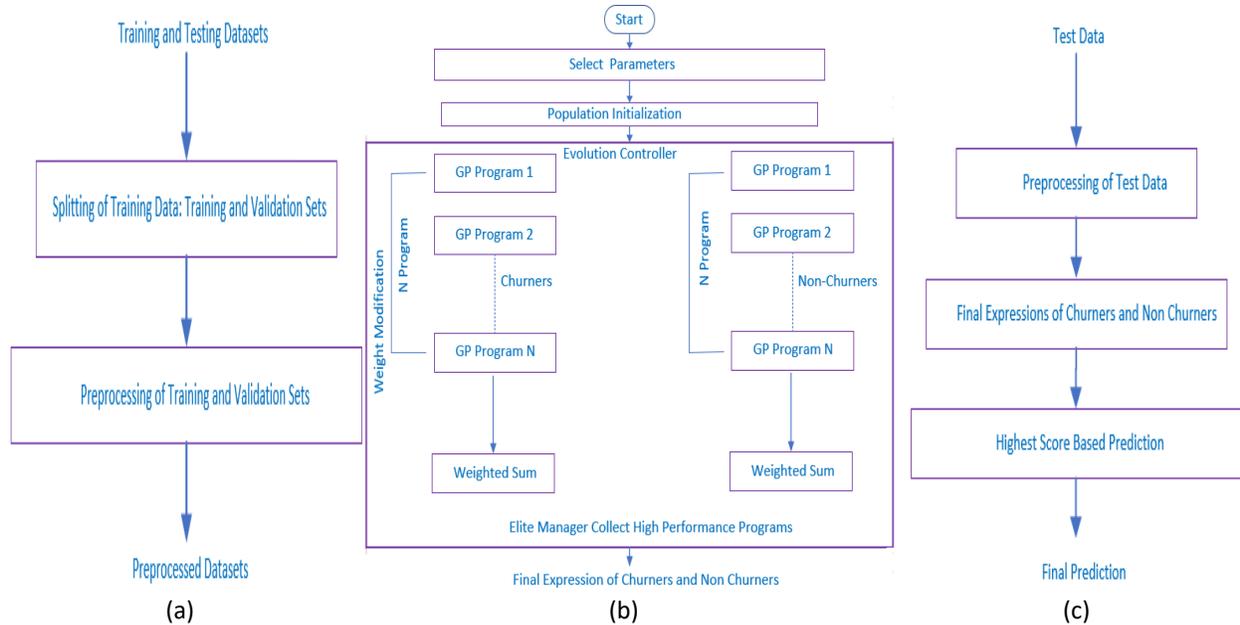

Figure 2: Block diagram of GP-AdaBoost approach: (a) Obtaining preprocessed datasets, (b) Developing the final expression for both Churners and non-Churners using GP-AdaBoost, and (c) Final churn prediction.

3.2 ROC Curve

The ROC curve and other related evaluation measurements are pertinent in two ways. Firstly, it provides a comparatively unbiased diagnostic ability of a binary classifier as compared to prediction accuracy, which can easily become biased in the case of imbalanced datasets. Secondly, it is helpful while comparing the performance of different methods, reported in the literature. Equations (2), and (3) show Sensitivity and Specificity parameters, while AUC is calculated as in (Equation 4) [35].

$$Sensitivity = \frac{TP}{TP+FN} \quad (2)$$

$$Specificity = \frac{TN}{TN+FP} \quad (3)$$

$$AUC = \int_0^1 \frac{TP}{TP+FN} \, d\, \frac{FP}{TN+FP} \tag{4}$$

## 4. Results and Discussion

Churn prediction system is assessed on the basis of its ability to correctly identify the churners in telecom data. At present, most of the reported churn prediction systems are not capable of achieving accurate predictions due to the serious complications in telecommunication data. The correct prediction of churners helps the telecom industry in avoiding a heavy loss. The proposed "TL-DeepE" method for churn prediction shows improved results for the standard telecom datasets.

### 4.1 Datasets

Two standard telecom datasets namely; Orange and Cell2cell are used for experimentation and analysis. The former one, i.e. Orange dataset from Telecom, UK, is made available to the researchers as part of KDD-Cup 2009 competition. The Cell2cell dataset, on the other hand, is from Duke University's Center for Customer Relationship Management. The Orange dataset has eighteen features with missing values and five features have just a single value, while Cell2cell dataset has no missing values. The symbolic feature values existing in telecommunication dataset are converted into numerical forms. Attributes of these telecommunication datasets are described in Table 1.

Table 1: Characteristics of Telecom Datasets

| Dataset | Cell2cell | Orange |
|---|---|---|
| **Source** | Duke University | KDD Cup 2009 |
| **Features** | 77 | 230 |
| **Samples** | 40000 | 50000 |
| **Features title** | Well- defined | Undefined |
| **Categorical variables** | 1 | 34 |
| **VisioBehavior** | Fair | Unfair |
| **Positive samples** | 20000 | 46328 |
| **Negative samples** | 20000 | 3672 |
| **Missing features value** | No | Yes |

### 4.2 Dataset Splits and Overall Working of the Prediction System

As regards the splitting of the data, we have used the hold-out method. However, in the meta-classification phase, we have used 10-fold cross-validation technique. The distribution of dataset must be taken in the account before the whole process starts because some portion of the dataset will be used for training and the rest in evaluating the GP-AdaBoost ensemble. Therefore, both the datasets are divided into two subsets A and B after applying proper pre-processing techniques. Neural network models are trained on the subset A, which encompasses 60% of the entire dataset. These CNNs are used as the trained base learner in the GP-AdaBoost ensemble. Once trained, the model predicts the outcome of their learning on the subset B, i.e. encompassing 40% of the total data. Once their test predictions on the subset B are completed, then these predictions are appended to the feature vector of subset B to form the extended feature vector and saved in subset B'. Finally, the performance of the GP-AdaBoost on subset B' is evaluated by using 10-fold cross-validation. The training and testing of the CNN models and the GP-AdaBoost classifier is shown in Figure 3. The final prediction is then provided by the GP-AdaBoost classifier.

### 4.3 Churn Prediction Using Individual CNN Classifiers and Transfer Learning

The performance of the base prediction models (pre-trained CNNs models that are fine-tuned) for Cell2cell and Orange datasets are summarized in Table 2. Inception-ResNet-V2 neural networks are pre-trained and then fine-tuned on telecom datasets. Transfer learning increases the prediction accuracies on Orange and Cell2cell to 63.57 and 68.01 %, respectively. It can be observed that the performance of the base models is not that high.

### 4.4 Ensemble's Performance without Transfer Learning

It was hypothesized that Transfer Learning can improve the overall churn prediction and classification performance. In order to validate this idea, it was essential to evaluate the performance of TL-DeepE without Transfer Learning. In this regard, Table 3 shows the GP-AdaBoost ensemble performance without Transfer Learning on Orange and Cell2cell

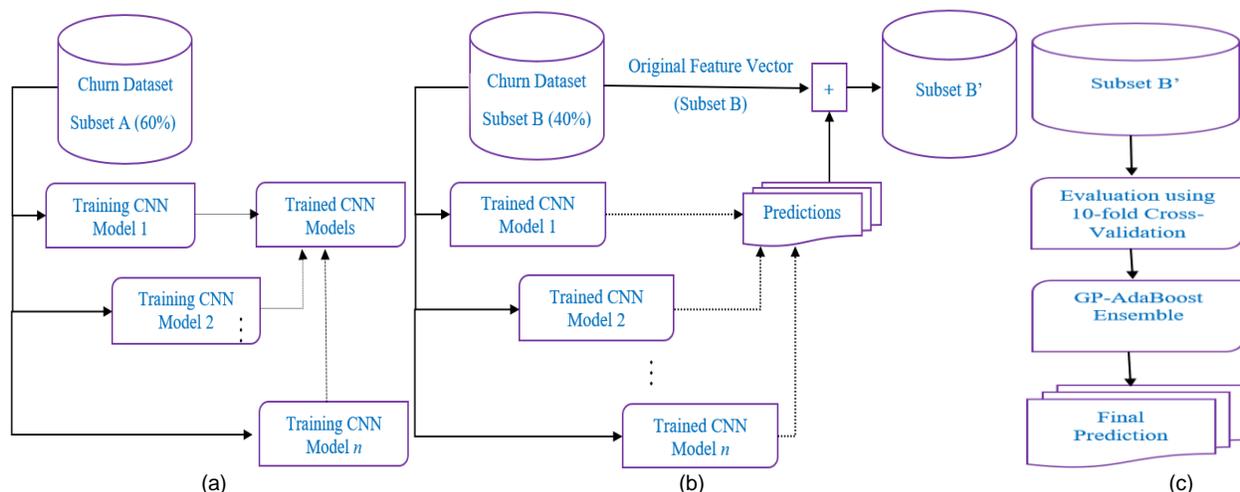
Figure 3: (a) Individual CNN Based Training, (b) Individual CNN Based Testing, (c) 10-fold GP-AdaBoost Based Training and Testing.

Table 2: Transfer Learning Based Results of Neural Network Models on Cell2cell and Orange Datasets.

| Runs | Prediction Accuracy (%) | | | | | |
|---|---|---|---|---|---|---|
| | Cell2cell | | | Orange | | |
| | AlexNet | Incep-RN-V2 | Custom-NN | AlexNet | Incep-RN-V2 | Custom-NN |
| Run 1 | 59.97 | 63.83 | 62.92 | 65.48 | 68.21 | 63.16 |
| Run 2 | 59.45 | 63.65 | 61.08 | 66.19 | 67.74 | 63.04 |
| Run 3 | 60.77 | 63.44 | 62.57 | 64.40 | 68.13 | 63.27 |
| Run 4 | 60.53 | 63.25 | 62.60 | 64.89 | 67.78 | 62.95 |
| Run 5 | 59.80 | 63.48 | 62.41 | 66.05 | 67.61 | 63.18 |
| Run 6 | 60.96 | 62.98 | 61.35 | 65.86 | 67.94 | 63.07 |
| Run 7 | 61.36 | 63.17 | 61.41 | 66.12 | 68.06 | 63.20 |
| Run 8 | 60.20 | 64.27 | 61.20 | 65.29 | 68.49 | 62.84 |
| Run 9 | 61.37 | 63.55 | 62.40 | 65.75 | 67.97 | 62.98 |
| Run 10 | 61.74 | 64.05 | 62.37 | 66.10 | 68.24 | 62.80 |
| **Average** | **60.62** | **63.57** | **62.03** | **65.61** | **68.01** | **63.05** |

datasets. It is observed that the performance of the GP-AdaBoost ensemble without Transfer learning is not satisfactory and even lower than some of the individual base learners.

### 4.5 Ensemble's Performance using Transfer Learning

Table 4 summarizes the results obtained by the proposed TL-DeepE using Transfer Learning. In this case, the performance evaluation of the high-level meta-classifier is performed on two standard telecommunication datasets using 10-fold cross-validation. The table depicts an average performance for 10 independent runs of the algorithm. Furthermore, AUC for Orange and Cell2cell datasets are 0.83 and 0.74, is shown in Figure 4(a). In Figure 4(b), comparisons are provided for GP-AdaBoost ensemble (using both with and without Transfer Learning), on Orange and Cell2cell datasets. It is observed that the performance of the GP-AdaBoost ensemble in combination with Transfer learning is boosted and becomes higher than all of the deep individual base learners.

### 4.6 Comparative Analysis

Performance of the proposed churn prediction technique, i.e. TL-DeepE, is compared with previous techniques, on Cell2cell and Orange telecom datasets. Initially, the performance of the churn predictors; AlexNet, ResNet-V2, Custom-CNN, and then GP-AdaBoost ensemble classifier without Transfer Learning is evaluated on the telecommunication dataset. The poor performance of Alex Net, ResNet-V2, Custom-CNN neural networks is because of the already performed pre-tuning and little fine-tuning on telecom datasets. Hence an effective prediction system is necessary in order to avoid customer churn related losses.

The proposed system "TL-DeepE" in (Figure 4 (b)) shows that the integration of information from Transfer Learning with GP-Adaboost ensemble improves churn prediction performance over the complex datasets and surpasses the performance of all the earlier reported systems. The proposed system achieved highest churn prediction accuracy, on Orange and Cell2cell telecommunication datasets, as 75.4% and 68.2%, while the area under the curve as 0.83 and 0.74, respectively. This shows its strength to perform on complicated nature of the telecom data.

Table 3: GP-AdaBoost Ensemble Performance on Cell2Cell and Orange Datasets without Transfer Learning.

| Runs | Cell2Cell | | Orange | |
|---|---|---|---|---|
| | Prediction Accuracy (%) | AUC | Prediction Accuracy (%) | AUC |
| Run 1 | 60.30 | 0.64 | 62.16 | 0.69 |
| Run 2 | 60.62 | 0.65 | 61.47 | 0.65 |
| Run 3 | 61.22 | 0.64 | 62.81 | 0.67 |
| Run 4 | 61.00 | 0.64 | 62.04 | 0.66 |
| Run 5 | 60.33 | 0.62 | 63.19 | 0.67 |
| Run 6 | 61.04 | 0.63 | 63.13 | 0.67 |
| Run 7 | 60.90 | 0.63 | 62.48 | 0.66 |
| Run 8 | 60.57 | 0.63 | 62.94 | 0.66 |
| Run 9 | 61.18 | 0.64 | 62.03 | 0.66 |
| Run 10 | 60.43 | 0.63 | 62.80 | 0.64 |
| **Average** | **60.76** | **0.64** | **62.50** | **0.66** |

Table 4: Prediction Performance of the proposed TL-DeepE on Cell2cell and Orange Datasets.

| Runs | Cell2Cell | | Orange | |
|---|---|---|---|---|
| | Prediction Accuracy (%) | AUC | Prediction Accuracy (%) | AUC |
| Run 1 | 68.07 | 0.74 | 76.13 | 0.83 |
| Run 2 | 67.86 | 0.74 | 75.90 | 0.83 |
| Run 3 | 69.30 | 0.75 | 75.33 | 0.83 |
| Run 4 | 68.65 | 0.74 | 75.51 | 0.83 |
| Run 5 | 68.42 | 0.75 | 74.94 | 0.82 |
| Run 6 | 67.38 | 0.73 | 75.11 | 0.83 |
| Run 7 | 67.74 | 0.73 | 75.77 | 0.83 |
| Run 8 | 68.40 | 0.74 | 75.50 | 0.83 |
| Run 9 | 67.91 | 0.73 | 74.11 | 0.83 |
| Run 10 | 67.84 | 0.73 | 75.51 | 0.83 |
| **Average** | **68.16** | **0.74** | **75.38** | **0.83** |

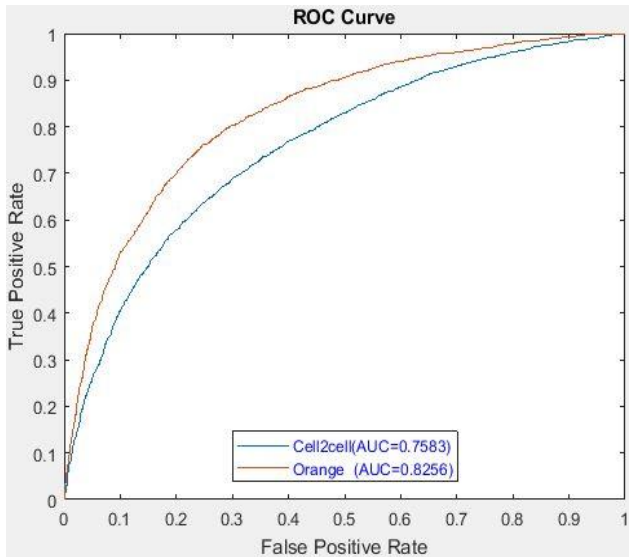
(a)

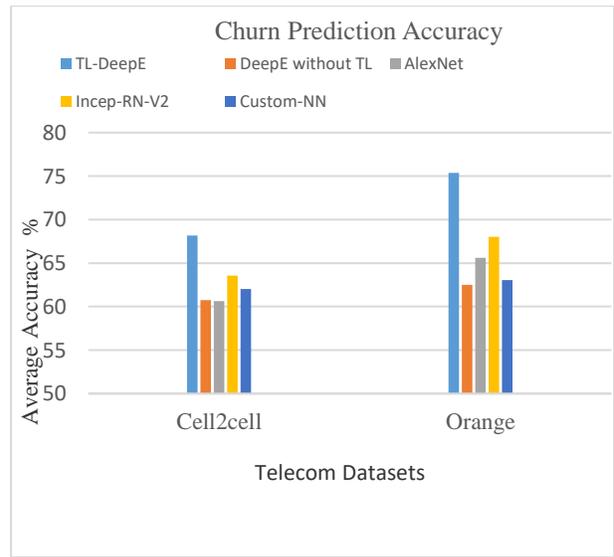
(b)

Figure 4. (a) ROC curve of GP- AdaBoost on Cell2cell and Orange dataset and (b) Comparison of Churn predictors.

## 5. Conclusion

A novel TL-DeepE technique is proposed to predict potential churners, which is very important for the competitive telecom industry. The size and dimensionality of the telecom data are high and require a large computational power for churn prediction. The proposed TL-DeepE makes use of multiple CNN architectures and GP-AdaBoost ensemble with the exploitation of Transfer Learning concepts. The individual prediction of the base original feature vectors of the dataset to form extended feature vectors. Finally, using a 10-fold cross-validation technique, a GP-AdaBoost ensemble as meta-classifier is evaluated to obtain predictions.

The proposed TL-DeepE churn prediction system has shown its effectiveness in predicting churners using standard telecommunication datasets. TL-DeepE has demonstrated a churn prediction accuracy of 75.4% and 0.83 AUC on Orange dataset. Its accuracy was 68.2% and AUC 0.74 on Cell2Cell dataset. It is observed that due to the challenging nature of the churn prediction task, choice of suitable features and the exploitation of Deep Learning, Transfer Learning, and GP-AdaBoost meta-classification appears more effective for modeling churn prediction for the telecommunication industry. The proposed method can be potentially effective in addressing the concerns and complications of the telecommunication industry.

### Acknowledgment:

This research was supported by HEC NRPU Research Grant (No. 20-3408/R&D/HEC/14/233) Pakistan, and Basic Science Research Program through the National Research Foundation of Korea (NRF) funded by the Ministry of Education (2017R1A2B2005065).